\documentclass[letterpaper]{article}
\usepackage{aaai}
\usepackage{times}
\usepackage{helvet}
\usepackage{courier}
\usepackage{amsmath,amssymb,amsfonts}
\usepackage{graphicx}
\usepackage{booktabs}
\usepackage{multirow}
\usepackage{algorithm}
\usepackage{algorithmic}

\title{Distilling Deep Reinforcement Learning into Interpretable Fuzzy Rules: An Explainable AI Framework}
\author{Sanup Araballi \\
Dept. of EECS,\\
Syracuse University, NY 13244, USA \And 
Simon Khan \\
Air Force Research Laboratory,\\
Rome, NY 13441, USA \And 
Chilukuri K Mohan \\
Dept. of EECS,\\
Syracuse University, NY 13244, USA 
}

\begin{document}
\maketitle

\begin{abstract}
\begin{quote}
Deep Reinforcement Learning (DRL) agents achieve remarkable performance in continuous control but remain opaque, hindering deployment in safety-critical domains. Existing explainability methods either provide only local insights (SHAP, LIME) or employ over-simplified surrogates failing to capture continuous dynamics (decision trees). This work proposes a Hierarchical Takagi-Sugeno-Kang (TSK) Fuzzy Classifier System (FCS) distilling neural policies into human-readable IF-THEN rules through K-Means clustering for state partitioning and Ridge Regression for local action inference. Three quantifiable metrics are introduced: Fuzzy Rule Activation Density (FRAD) measuring explanation focus, Fuzzy Set Coverage (FSC) validating vocabulary completeness, and Action Space Granularity (ASG) assessing control mode diversity. Dynamic Time Warping (DTW) validates temporal behavioral fidelity. Empirical evaluation on \textit{Lunar Lander(Continuous)} shows the Triangular membership function variant achieves 81.48\% $\pm$ 0.43\% fidelity, outperforming Decision Trees by 21 percentage points. The framework exhibits statistically superior interpretability (FRAD = 0.814 vs. 0.723 for Gaussian, $p < 0.001$) with low MSE (0.0053) and DTW distance (1.05). Extracted rules such as ``IF lander drifting left at high altitude THEN apply upward thrust with rightward correction'' enable human verification, establishing a pathway toward trustworthy autonomous systems.    
\end{quote}
\end{abstract}

\section{Introduction}

Reinforcement Learning (RL) has transformed sequential decision-making, enabling agents to master complex tasks ranging from robotic manipulation \cite{levine2016end} to autonomous navigation \cite{kahn2018self} and game playing \cite{silver2016mastering,mnih2015human}. Algorithms such as Proximal Policy Optimization (PPO) \cite{schulman2017proximal} allow agents to learn sophisticated policies directly from high-dimensional sensory inputs. However, these policies are encoded within the weights of deep neural networks. For a human observer, such as a safety engineer or regulatory authority, understanding why an agent selected a specific continuous action remains difficult. For example, if a spacecraft landing agent applies a thrust vector of $[0.36, 0.71]$ or torque of $[-0.37, 0.02]$, the underlying logic driving this decision is not explicitly available from the network parameters. This opacity presents a barrier to deployment in domains where verification, certification, and accountability are essential \cite{gunning2019explainable}.

Consider an autonomous lunar landing agent trained via PPO. The policy network contains multiple hidden layers with hundreds of parameters, each contributing in a non-linear manner to the final action. When the agent reduces main engine thrust to $-0.37$ while applying side thrust of $0.02$, the decision rationale is inaccessible. Was this based on the lander's tilt angle? A complex interaction between position and velocity? The neural network provides no answer. This lack of transparency limits trust, prevents failure diagnosis, makes formal verification and safety certifications impossible.

The field of Explainable Reinforcement Learning (XRL) \cite{puiutta2020explainable,wells2020explainable} aims to bridge this gap. However, explainability in continuous control differs substantially from discrete classification tasks. Continuous control requires approximating smooth, non-linear mappings from state spaces to action spaces. Current XRL approaches fall into three categories, each with significant limitations:

\textbf{Local Explanation Methods.} Techniques such as SHAP (SHapley Additive exPlanations) \cite{lundberg2017unified} and LIME (Local Interpretable Model-agnostic Explanations) \cite{ribeiro2016should} provide feature attribution for individual states \cite{greydanus2018visualizing}. These methods answer ``which features contributed to this action?'' but do not reveal global policy structure or operational modes. A practitioner examining SHAP values for many states gains no holistic understanding. The explanations are instance-specific and not portable. As noted by Rudin \cite{rudin2019stop}, local explanations can mislead if not globally consistent---a particular concern in RL where temporal dependencies can produce contradictory explanations across states.

\textbf{Symbolic Distillation Approaches.} Decision trees offer global interpretability. VIPER \cite{bastani2018verifiable} trains trees via imitation learning to mimic Q-networks. However, trees partition state spaces into regions with constant action outputs. To approximate a smooth control function, a tree must partition the state space into many fine-grained regions. This often results in trees that are either too deep to be interpretable or too shallow to be accurate. Each leaf outputs a fixed action, causing abrupt discontinuities at region boundaries. Silva et al. \cite{silva2020optimization} explored soft decision trees to improve smoothness, but core discretization issues persist. Program synthesis \cite{verma2018programmatically} and symbolic regression \cite{landajuela2022discovering} discover closed-form expressions but face severe scalability challenges in high-dimensional spaces.

\textbf{Neuro-Fuzzy Methods.} Fuzzy Logic Systems \cite{zadeh1965fuzzy} naturally bridge numerical computation and linguistic reasoning. By representing continuous variables with linguistic terms (e.g., ``Velocity is High'') and combining them via IF-THEN rules, fuzzy systems express complex relationships in human-readable formats. Takagi-Sugeno-Kang (TSK) systems \cite{takagi1985fuzzy} output continuous functions of inputs rather than discrete fuzzy sets, allowing smooth interpolative predictions well-suited for control. Fuzzy Q-Learning \cite{glorennec1994fuzzy} and Neuro-Fuzzy Actor-Critic methods \cite{jouffe1998fuzzy} integrate fuzzy logic into RL training. While interpretable by design, these approaches underperform modern deep RL on complex tasks. Post-hoc fuzzy surrogates \cite{pan2019fuzzy,gu2020neuro} show promise but lack: (1) hierarchical strategies for scalability, (2) rigorous quantifiable metrics beyond fidelity, and (3) comprehensive validation against baselines.

To overcome these limitations, this work proposes the below mentioned contributions from our work.

\subsection{Contributions}

The contributions of this work are fourfold:
\begin{enumerate}
    \item A Hierarchical TSK Fuzzy Classifier System is proposed that effectively decouples state partitioning from action inference, enabling scalability to continuous control tasks.
    \item Gaussian and Triangular membership functions are systematically compared, demonstrating that Triangular functions yield more focused and interpretable explanations without sacrificing fidelity.
    \item Three mathematically grounded metrics (FRAD, FSC, ASG) are introduced to rigorously quantify the quality of rule-based explanations.
    \item A comprehensive empirical evaluation is provided, showing that the fuzzy surrogate achieves superior balance of fidelity and interpretability compared to symbolic baselines: 81.5\% fidelity versus 60.1\% for decision trees.
\end{enumerate}

\section{Related Work}

Explainability in RL intersects multiple research areas: interpretable machine learning, fuzzy logic, symbolic AI, and formal verification. This section reviews prior work, highlighting achievements and limitations motivating this framework.

\subsection{Explainable Reinforcement Learning}

The field of Explainable AI (XAI) has grown substantially, driven by regulatory requirements and deployment needs in high-stakes domains \cite{gunning2019explainable,arrieta2020explainable}. Adapting XAI to RL is challenging due to sequential decision-making where actions have delayed consequences and policies couple to value functions estimating future rewards.

\textbf{Saliency and Attribution Methods.} Greydanus et al. \cite{greydanus2018visualizing} applied saliency maps to Atari agents, visualizing which pixels influenced actions. Zahavy et al. \cite{zahavy2016graying} used perturbation-based techniques for feature importance. SHAP \cite{lundberg2017unified}, providing Shapley value-based attributions from game theory, has been adapted to RL contexts \cite{bekkemoen2021shap}. While these methods offer valuable local insights, they remain limited in scope. They do not reveal global policy structure or operational modes. As argued by Rudin \cite{rudin2019stop}, local explanations can be misleading if not globally consistent---a particular concern in RL where temporal dependencies can produce contradictory explanations across states.

\textbf{Policy Summarization.} Alternative approaches provide explanations through selected trajectories. Amir and Amir \cite{amir2018highlights} proposed extracting trajectory ``highlights'' that explain strategy. Huang et al. \cite{huang2018establishing} developed methods for generating concise summaries conveying policy intent. While trajectory-based explanations can be intuitive, they remain exemplar-based: they show what the agent does in specific scenarios but do not provide a generalizable model of how decisions are made across the state space.

\subsection{Symbolic Policy Distillation}

Global interpretability requires distilling policies into transparent symbolic representations. The VIPER framework \cite{bastani2018verifiable} trains decision trees via imitation learning to mimic Q-networks. Trees partition state spaces into hyper-rectangles with constant action outputs. This creates piecewise-constant approximations of smooth control functions. Achieving acceptable fidelity requires extremely deep trees, sacrificing interpretability. Silva et al. \cite{silva2020optimization} explored soft decision trees and differentiable architectures to improve smoothness, but core discretization issues remain. Decision trees are axis-aligned and struggle with non-separable decision boundaries common in RL state spaces.

Linear models \cite{liu2018towards} are maximally interpretable but typically too simple for complex non-linear policies. Program synthesis \cite{verma2018programmatically} and symbolic regression \cite{landajuela2022discovering} attempt to discover closed-form mathematical expressions replicating policies. These methods face severe scalability challenges in high-dimensional spaces and are often limited to toy problems.

Learning Classifier Systems (LCS), particularly XCS \cite{wilson1995classifier,butz2002xcs}, learn rule populations via genetic algorithms and are inherently interpretable. Traditional LCS methods target discrete action spaces and low-dimensional states. Extensions like XCSF \cite{lanzi2005xcs,wilson2002classifiers} handle continuous inputs but face scalability challenges and have not been widely adopted for explaining modern deep RL policies.

\subsection{Fuzzy Systems in Reinforcement Learning}

Fuzzy Logic Systems \cite{zadeh1965fuzzy} bridge numerical computation and linguistic reasoning through fuzzification, rule inference, and defuzzification. TSK systems \cite{takagi1985fuzzy} eliminate defuzzification by directly outputting continuous functions (typically linear: $f_i(\mathbf{x}) = \mathbf{w}_i^T \mathbf{x} + b_i$), providing smooth interpolation ideal for control.

\textbf{Fuzzy RL Algorithms.} Fuzzy logic has been integrated into RL training loops. Glorennec \cite{glorennec1994fuzzy} introduced Fuzzy Q-Learning associating Q-values with fuzzy state-action pairs. Jouffe \cite{jouffe1998fuzzy} developed neuro-fuzzy actor-critic architectures. Hein and Udluft \cite{hein2008continuous} proposed evolving fuzzy RL for continuous domains. While interpretable by design, these approaches face scalability limitations and performance gaps compared to modern deep RL algorithms like PPO, SAC, and TD3.

\textbf{Post-Hoc Fuzzy Surrogates.} More aligned with this work is the use of fuzzy systems as post-hoc surrogates explaining pre-trained neural policies. Pan et al. \cite{pan2019fuzzy} extracted TSK rules from deep belief networks for classification tasks. Gu et al. \cite{gu2020neuro} explored neuro-fuzzy systems for explainable AI on supervised learning. In RL, Abels et al. \cite{abels2019dynamic} used fuzzy cognitive maps for visualization but did not provide global surrogates. Hayes and Shah \cite{hayes2017improving} extracted fuzzy rules from discrete-action Q-networks.

This work advances the state-of-the-art through: (1) a hierarchical learning strategy scaling fuzzy surrogates to continuous control by decoupling state partitioning from action inference, (2) rigorous comparison of Gaussian vs. Triangular membership functions demonstrating interpretability advantages, and (3) introduction of quantifiable metrics (FRAD, FSC, ASG) enabling objective evaluation of explanation quality.

\section{Methodology}

The Hierarchical Takagi-Sugeno-Kang (TSK) Fuzzy Classifier System is presented in this section which will act as a surrogate to the trained neural network. We further present a formal problem formulation, two-level architecture, and explainable AI metric definitions.

\subsection{Problem Formulation}

A continuous control RL setting is considered, defined by MDP $\mathcal{M} = (\mathcal{S}, \mathcal{A}, P, r, \gamma)$, where $\mathcal{S} \subseteq \mathbb{R}^d$ is the state space, $\mathcal{A} \subseteq \mathbb{R}^m$ is the action space, $P$ is the transition function, $r$ is the reward function, and $\gamma$ is the discount factor. A neural policy $\pi_{\theta}$ maps states to actions. The goal is constructing surrogate $\pi_{\text{FCS}}$ that: (1) accurately approximates pre-trained teacher $\pi_{\text{NN}}$, (2) is structured as human-readable IF-THEN rules, and (3) satisfies quantifiable explainability metrics. A supervised imitation learning framework is adopted, collecting dataset $\mathcal{D} = \{(\mathbf{s}_i, \mathbf{a}_i)\}$ from teacher rollouts.

\subsection{Hierarchical Architecture}

\begin{figure}[!t]
\centering
\includegraphics[width=\columnwidth]{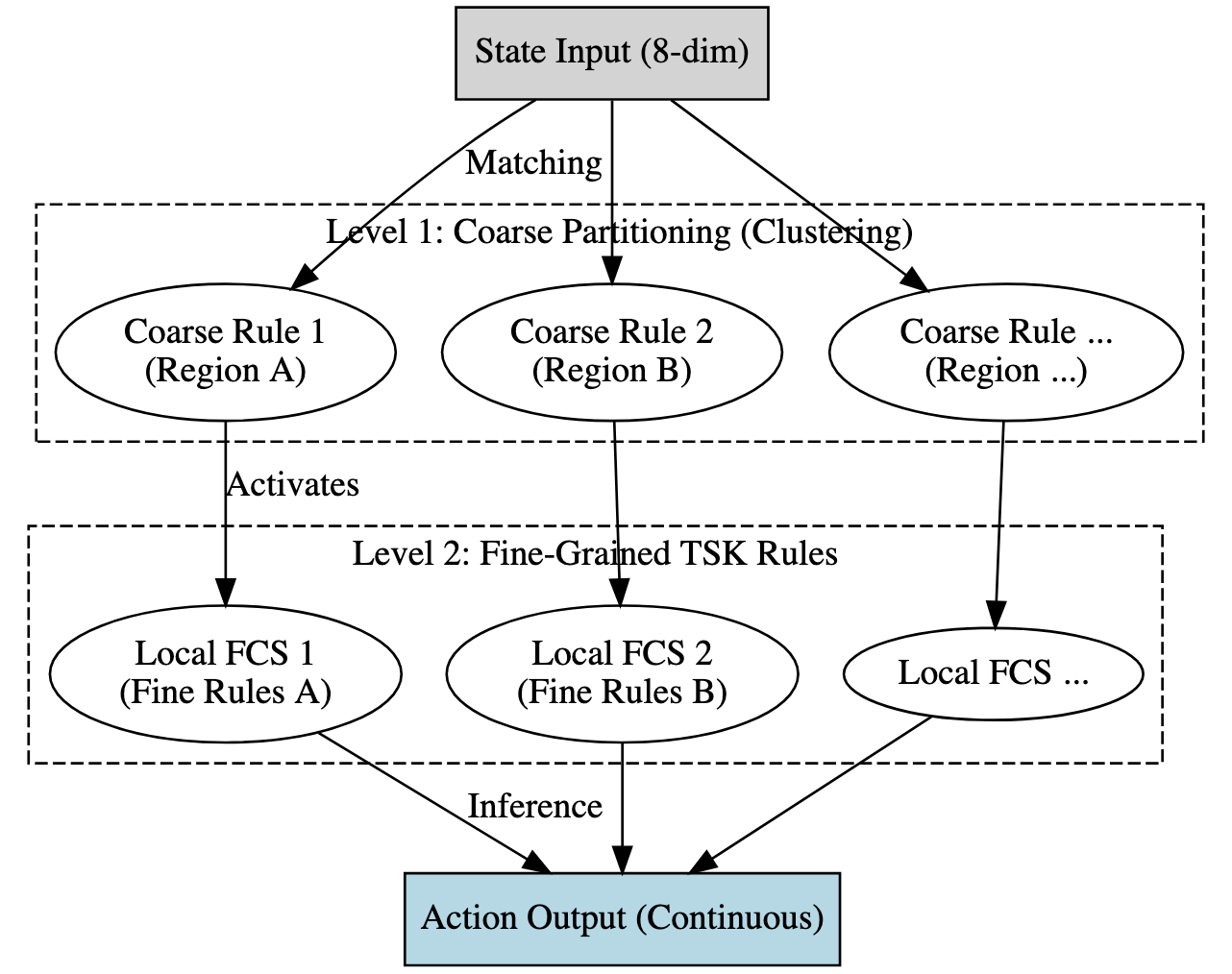}
\caption{Hierarchical TSK FCS Architecture. Level 1 partitions state space via K-Means clustering into operational regions. Level 2 learns local TSK consequents within each region via Ridge Regression. Final actions obtained through normalized weighted aggregation of active local models.}
\label{fig:architecture}
\end{figure}

A two-level hierarchical decomposition is employed to address scalability challenges. Figure~\ref{fig:architecture} illustrates the architecture.

\textbf{Level 1: Antecedent Learning.} K-Means clustering \cite{macqueen1967some} partitions state space $\mathcal{S}$ into $N$ regions. Each cluster $i$ with centroid $\mathbf{c}_i$ and spread $\sigma_i$ represents an operational mode (e.g., ``hovering,'' ``correcting drift''). For each cluster $i$ and state dimension $k$, a membership function $\mu_{i,k}: \mathbb{R} \rightarrow [0,1]$ is defined. Two types are investigated:

\textit{Gaussian Membership Functions:}
\begin{equation}
\mu_{i,k}^{\text{Gauss}}(s_k) = \exp\left( - \frac{(s_k - c_{i,k})^2}{2\sigma_{i,k}^2} \right)
\label{eq:gaussian_mf}
\end{equation}
Gaussian functions provide smooth transitions with infinite support. Every state has non-zero membership in all clusters.

\textit{Triangular Membership Functions:}
\begin{equation}
\mu_{i,k}^{\text{Tri}}(s_k) = \max\left(0, \min\left( \frac{s_k - l_{i,k}}{c_{i,k} - l_{i,k}}, \frac{r_{i,k} - s_k}{r_{i,k} - c_{i,k}} \right) \right)
\label{eq:triangular_mf}
\end{equation}
where $l_{i,k} = c_{i,k} - \beta \sigma_{i,k}$, $r_{i,k} = c_{i,k} + \beta \sigma_{i,k}$ (typically $\beta = 1.5$). Triangular functions offer compact support: membership is zero outside interval $[l_{i,k}, r_{i,k}]$.

The design choice between Gaussian and Triangular functions represents a trade-off. Gaussian functions provide smooth interpolation across all states but activate many rules simultaneously, potentially diluting explanation clarity. Triangular functions create localized activations where fewer rules are active, making decision logic more transparent. The hypothesis is that Triangular functions yield higher FRAD (more focused explanations) without sacrificing fidelity.

The firing strength of rule $i$ for state $\mathbf{s}$ is:
\begin{equation}
\alpha_i(\mathbf{s}) = \prod_{k=1}^{d} \mu_{i,k}(s_k)
\label{eq:firing_strength}
\end{equation}
For Triangular functions, if any dimension has zero membership, the entire rule's activation becomes zero, so that the rule sets are sparse and facilitate human comprehension.

\textbf{Level 2: Consequent Learning.} Within each region $i$, local control dynamics are modeled using a TSK consequent function. Unlike Mamdani systems requiring defuzzification, TSK rules directly output crisp functions:
\begin{equation}
f_i(\mathbf{s}) = \mathbf{w}_i^T \mathbf{s} + b_i
\label{eq:tsk_consequent}
\end{equation}
where $\mathbf{w}_i \in \mathbb{R}^d$ and $b_i \in \mathbb{R}$. This linear formulation balances expressiveness and interpretability: $\mathbf{w}_i$ captures how actions vary with state features; $b_i$ represents the region's baseline action.

Parameters are learned via weighted Ridge Regression \cite{hoerl1970ridge}. For each rule $i$, samples $(\mathbf{s}_j, a_j) \in \mathcal{D}$ are weighted by firing strength $\alpha_i(\mathbf{s}_j)$:
\begin{equation}
\min_{\mathbf{w}_i, b_i} \sum_{j=1}^{N_{\text{data}}} \alpha_i(\mathbf{s}_j) \left( a_j - (\mathbf{w}_i^T \mathbf{s}_j + b_i) \right)^2 + \lambda \|\mathbf{w}_i\|^2
\label{eq:ridge_regression}
\end{equation}
where $\lambda$ is the regularization parameter. Ridge regularization prevents overfitting when rules have limited support in data.

\textbf{Global Inference.} For a new state $\mathbf{s}$, the final action is:
\begin{equation}
a_{\text{FCS}}(\mathbf{s}) = \frac{\sum_{i=1}^{N} \alpha_i(\mathbf{s}) \cdot f_i(\mathbf{s})}{\sum_{i=1}^{N} \alpha_i(\mathbf{s})}
\label{eq:global_inference}
\end{equation}
This standard TSK formula provides smooth interpolation: states near centroids are primarily influenced by corresponding rules; boundary states blend multiple rules proportionally. For Triangular functions, $\alpha_i(\mathbf{s}) = 0$ for many rules, leading to sparse, localized inference.

\subsection{Novel Explainability Measures}

Explainability must be quantifiable for rigorous scientific evaluation. The three measures, Fuzzy Rule Activation Density, Fuzzy Set Coverage and Action Space Granularity, are formally defined capturing orthogonal aspects of explanation quality.

\textbf{Fuzzy Rule Activation Density (FRAD)} \newline
\textit{Motivation:} An explanation is most useful when decisive rather than ambiguous. If many rules activate with similar strengths, decision logic is unclear. If a single rule dominates, the explanation is focused.

\textit{Definition:} FRAD is inspired by the Herfindahl-Hirschman Index \cite{hirschman1964paternity}, measuring market concentration in economics. For state $\mathbf{s}$ with firing strengths $\{\alpha_i(\mathbf{s})\}_{i=1}^{N}$:
\begin{equation}
\text{FRAD}(\mathbf{s}) = \sum_{i=1}^{N} \left( \frac{\alpha_i(\mathbf{s})}{\sum_{j=1}^{N} \alpha_j(\mathbf{s})} \right)^2
\label{eq:frad}
\end{equation}

\textit{Interpretation:} FRAD ranges from $1/N$ (all rules fire equally, maximum ambiguity) to 1.0 (single rule fires, maximum focus). Higher values indicate more focused explanations. Mean FRAD over dataset $\mathcal{D}_{\text{test}}$:
\begin{equation}
\overline{\text{FRAD}} = \frac{1}{|\mathcal{D}_{\text{test}}|} \sum_{\mathbf{s} \in \mathcal{D}_{\text{test}}} \text{FRAD}(\mathbf{s})
\label{eq:mean_frad}
\end{equation}

\textbf{Fuzzy Set Coverage (FSC)} \newline
\textit{Motivation:} For valid explanations, linguistic vocabulary (fuzzy membership functions) must adequately cover the operational domain. If regions exist where all fuzzy sets have low membership, the surrogate must extrapolate, making explanations unreliable.

\textit{Definition:} For state $\mathbf{s} = (s_1, \ldots, s_d)$ and dimension $k$, the maximum membership across rules is:
\begin{equation}
\text{MaxMem}_k(\mathbf{s}) = \max_{i \in \{1, \ldots, N\}} \mu_{i,k}(s_k)
\label{eq:maxmem}
\end{equation}
FSC averages maximum membership across dimensions and states:
\begin{equation}
\text{FSC} = \frac{1}{|\mathcal{D}_{\text{test}}|} \sum_{\mathbf{s} \in \mathcal{D}_{\text{test}}} \left( \frac{1}{d} \sum_{k=1}^{d} \text{MaxMem}_k(\mathbf{s}) \right)
\label{eq:fsc}
\end{equation}

\textit{Interpretation:} FSC $\approx$ 1.0 indicates good coverage with at least one fuzzy set having high membership for most states and dimensions. Low FSC reveals gaps requiring extrapolation.

\textbf{Action Space Granularity (ASG)} \newline
\textit{Motivation:} Good explanations reveal distinct behavioral modes. If all rule consequents produce similar outputs, the rules fail to capture control strategy diversity.

\textit{Definition:} ASG measures variance of bias terms across rules:
\begin{equation}
\text{ASG} = \text{Var}(\{b_1, b_2, \ldots, b_N\}) = \frac{1}{N} \sum_{i=1}^{N} (b_i - \bar{b})^2
\label{eq:asg}
\end{equation}

\textit{Interpretation:} High ASG indicates rules represent qualitatively different actions (e.g., ``Thrust Up,'' ``Thrust Down,'' ``Hover''). Low ASG suggests rule collapse with similar outputs.

\textbf{Dynamic Time Warping for Behavioral Fidelity} \newline
\textit{Motivation:} MSE measures point-wise error but ignores temporal dynamics. In sequential tasks, behavioral similarity over trajectories is more important than instantaneous accuracy.

\textit{Definition:} Dynamic Time Warping (DTW) \cite{sakoe1978dynamic} compares time series, accounting for temporal shifts. For trajectories $\tau_{\text{NN}} = \{(\mathbf{s}_t^{\text{NN}}, \mathbf{a}_t^{\text{NN}})\}_{t=1}^{T}$ and $\tau_{\text{FCS}} = \{(\mathbf{s}_t^{\text{FCS}}, \mathbf{a}_t^{\text{FCS}})\}_{t=1}^{T'}$ initialized from identical states:
\begin{equation}
D(\tau_{\text{NN}}, \tau_{\text{FCS}}) = \text{DTW}(\tau_{\text{NN}}, \tau_{\text{FCS}})
\label{eq:dtw}
\end{equation}
Lower DTW indicates higher temporal similarity, validating that surrogates replicate control strategies over time.

\section{Experimental Setup}

An empirical study was designed addressing three research questions: (RQ1) Can hierarchical fuzzy surrogates achieve competitive fidelity while maintaining interpretability? (RQ2) Do Triangular functions yield more interpretable explanations than Gaussian? (RQ3) Can extracted rules provide semantically meaningful insights?

\subsection{Environment, Teacher, and Data}

The \texttt{LunarLanderContinuous-v3} environment \cite{brockman2016openai} was used as testbed. The state space ($\mathcal{S} \subseteq \mathbb{R}^8$) includes position, velocities, angle, angular velocity, and leg contacts. Actions ($\mathcal{A} \subseteq [-1, 1]^2$) control main engine thrust and side thrusters.

A PPO teacher \cite{schulman2017proximal} was trained via Stable-Baselines3 \cite{raffin2021stable} using a two-layer MLP (64 units/layer, tanh) for 50,000 timesteps, achieving stable performance (rewards $>$ 200). A dataset of 5,000 state-action pairs was collected via teacher rollouts, split into 80\% training and 20\% validation.

\subsection{Baselines and Configurations}

\textbf{Baselines:} (1) Decision Tree (CART \cite{pedregosa2011scikit}, max 16 leaves), and (2) Simple MLP (32 hidden units, ReLU).

\textbf{FCS Variants:} FCS-Gaussian-16, FCS-Triangular-16, FCS-Triangular-4, FCS-Triangular-8. K-Means with default settings; Triangular $\beta = 1.5$; Ridge $\lambda = 0.1$.

\subsection{Evaluation}

All experiments repeated across 5 random seeds (42-46). Metrics include: Action-Matching Accuracy (\%), MSE, DTW Distance (mean over 10 rollouts), FRAD, FSC, and ASG. Statistical significance assessed via paired t-tests ($\alpha = 0.05$). Implementation: Python 3.10, PyTorch 1.13, scikit-learn 1.2.

\section{Results}

The empirical evaluation validates the hierarchical fuzzy framework across multiple dimensions. Quantitative results demonstrate that the FCS-Triangular model achieves a superior balance of fidelity and interpretability compared to both symbolic and black-box baselines. The FCS-Triangular-4 (rules) variant exhibits unexpectedly high fidelity, revealing that complex neural policies can be distilled into remarkably simple structures. Extracted rules provide concrete evidence of genuine explainability, with semantic interpretations directly corresponding to operational control strategies. Statistical tests confirm key hypotheses regarding membership function design choices. The following subsections present detailed findings organized by evaluation dimension.

\subsection{Quantitative Comparison}

Table~\ref{tab:main_results} summarizes performance across all models (mean $\pm$ std over 5 seeds).

\begin{table}[!t]
\centering
\caption{Performance Comparison of Surrogate Models. Higher is better for Fidelity and FRAD; lower is better for MSE and DTW.}
\label{tab:main_results}
\scriptsize
\begin{tabular}{lccccc}
\toprule
\textbf{Model} & \textbf{Fidelity} & \textbf{MSE} & \textbf{DTW} & \textbf{FRAD} & \textbf{FSC} \\
 & \textbf{(\%)} & & & & \\
\midrule
Simple-MLP & \textbf{96.84} & \textbf{0.0016} & \textbf{0.55} & N/A & N/A \\
 & $\pm$1.80 & $\pm$0.001 & $\pm$0.20 & & \\
\midrule
\textbf{FCS-Tri.} & \textbf{81.48} & 0.0053 & 1.05 & \textbf{0.814} & 0.933 \\
 & $\pm$\textbf{0.43} & $\pm$0.001 & $\pm$0.17 & $\pm$\textbf{0.009} & $\pm$0.002 \\
FCS-Gaus. & 81.38 & \textbf{0.0037} & \textbf{0.87} & 0.723 & \textbf{0.974} \\
 & $\pm$0.64 & $\pm$0.000 & $\pm$0.05 & $\pm$0.010 & $\pm$0.001 \\
\midrule
DT-16 & 60.14 & 0.0074 & 1.32 & N/A & N/A \\
 & $\pm$1.27 & $\pm$0.001 & $\pm$0.14 & & \\
\bottomrule
\end{tabular}
\end{table}

\textbf{Key Findings:}

\textbf{Finding 1: Fuzzy Surrogates Bridge the Gap.} FCS-Triangular achieves 81.48\% $\pm$ 0.43\% fidelity, representing a 21.34 percentage point improvement over Decision Tree (60.14\%). This validates that fuzzy interpolation via TSK consequents is superior to piecewise-constant approximations. The Decision Tree, despite having 16 leaves matching the FCS rule count, cannot capture smooth action modulation. In contrast, the FCS learns local linear functions enabling smooth transitions. The FCS achieves high fidelity while maintaining a compact, rule-based structure. Each rule can be written as a human-readable IF-THEN statement (examples in Section V-E). This transparency is absent in both MLP and neural teacher.

\textbf{Finding 2: Triangular Functions Yield More Focused Explanations.} FCS-Gaussian and FCS-Triangular achieve similar fidelity (81.38\% vs. 81.48\%, $p > 0.05$). However, they differ significantly in interpretability focus. FCS-Triangular achieves FRAD of 0.814 $\pm$ 0.009 compared to 0.723 $\pm$ 0.010 for Gaussian. Paired t-test yields $t = 14.5$, $p < 0.001$ (highly significant). Higher FRAD indicates fewer simultaneously active rules. For a typical state, Triangular FCS activates fewer rules with concentrated activation mass. This makes decision logic more decisive. A domain expert examining explanations encounters fewer rules to consider, making rationale clearer.

\textbf{Finding 3: FSC and ASG Validate Quality.} FSC values are high (0.933 for Triangular, 0.974 for Gaussian), indicating fuzzy sets are well-positioned to cover the operational domain. Gaussian achieves slightly higher FSC due to infinite support. Both models exhibit excellent coverage, ensuring surrogates don't rely on unsafe extrapolation.

ASG values (0.173 for Triangular, 0.126 for Gaussian) confirm learned rules represent diverse control modes. Variance in bias terms indicates different rules encode distinct actions: strong upward thrust, gentle hovering, corrective maneuvers. This diversity is essential for informative rather than degenerate explanations.

\textbf{Finding 4: MLP Upper Bound.} Simple-MLP achieves 96.84\% fidelity, as expected for a flexible non-linear approximator. However, this comes at complete cost of interpretability. The MLP confirms the action prediction task is learnable and establishes the fidelity ceiling. FCS achieving 81\% with full interpretability represents a useful compromise.

\subsection{Rule Complexity Trade-off}

The impact of rule count on performance was investigated by training FCS-Triangular with $N \in \{4, 8, 16\}$. Table~\ref{tab:rule_complexity} presents results.

\begin{table}[!t]
\centering
\caption{Impact of Rule Count on FCS-Triangular Performance.}
\label{tab:rule_complexity}
\scriptsize
\begin{tabular}{lcccc}
\toprule
\textbf{\#Rules} & \textbf{Fidelity (\%)} & \textbf{MSE} & \textbf{FRAD} & \textbf{FSC} \\
\midrule
4  & \textbf{97.83 $\pm$ 0.5} & \textbf{0.00069} & 0.863 & 0.937 \\
8  & 95.83 $\pm$ 0.7 & 0.00119 & 0.801 & 0.963 \\
16 & 81.48 $\pm$ 0.4 & 0.00534 & \textbf{0.814} & 0.933 \\
\bottomrule
\end{tabular}
\end{table}

\textbf{Observation: The ``Less is More'' Phenomenon.} Counterintuitively, the 4-rule model achieves highest fidelity (97.83\%), outperforming the 16-rule model. This suggests the teacher policy's decision logic is governed by a small number of dominant modes. Forcing 16 rules may over-partition the state space, fitting noise or fragmenting coherent modes. This finding indicates the framework can identify minimal rule sets required for core strategy explanation, avoiding unnecessary complexity.

However, interpretability is not solely about minimizing rules. The 16-rule model may provide more granular insights into edge cases. The choice of $N$ should be guided by application: minimal rule sets (4-8) for deployment verification, larger sets (16+) for detailed debugging.

\subsection{Behavioral Fidelity via DTW}

\begin{figure}[!t]
\centering
\includegraphics[width=\columnwidth]{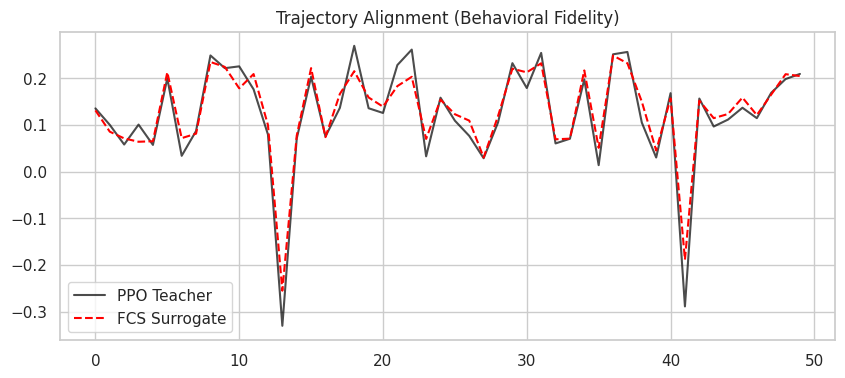}
\caption{Trajectory alignment: PPO teacher (black solid) vs. FCS surrogate (red dashed) over 50 timesteps. Close tracking demonstrates high behavioral fidelity (DTW = 1.03).}
\label{fig:trajectory_alignment}
\end{figure}

To validate that surrogates replicate temporal control strategy beyond static action matching, both policies were rolled out from 10 identical initial states. DTW distances were computed over resulting trajectories. Figure~\ref{fig:trajectory_alignment} shows a representative comparison.

The FCS surrogate (red dashed) closely tracks the teacher trajectory (black solid). Low DTW distance (mean 1.05 $\pm$ 0.17) confirms the surrogate replicates the temporal control strategy, including corrective maneuvers and stabilization behaviors. This validation is critical: in safety-critical applications, temporal consistency matters. An agent maintaining qualitatively similar trajectories is more trustworthy than one with low MSE but divergent behavior. DTW provides this holistic assessment.

\subsection{Interpretability Focus: FRAD Distributions}



Comparatively, the Triangular membership function yields higher FRAD of 0.814 vs. Gaussian membership function's 0.723 ($p < 0.001$), indicating more focused explanations.

This confirms the hypothesis: compact-support membership functions create sparser rule activations, making decision logic more transparent. For human auditors, this translates to fewer rules requiring consideration when examining rationale.

\subsection{Extracted Rules and Semantic Interpretability}

To demonstrate practical utility beyond aggregate statistics, concrete examples of human-readable fuzzy rules are presented from FCS-Triangular-16. Each rule is expressed in standard IF-THEN format with linguistic labels. Five representative rules are shown (out of 16 total).

\textbf{Rule 1: Stabilization from Leftward Drift}
\begin{quote}
\small
\textbf{IF} $X$ is NEG ($\sim -0.61$) \textbf{AND} $Y$ is POS ($\sim 1.98$) \\
\textbf{AND} $V_x$ is NEG ($\sim -0.37$) \textbf{AND} $V_y$ is ZERO ($\sim 0.12$) \\
\textbf{AND} $\theta$ is ZERO ($\sim -0.03$) \textbf{AND} $\omega$ is ZERO ($\sim -0.09$) \\
\textbf{THEN} Action is [Main = 0.36, Side = 0.71]
\end{quote}
\textit{Interpretation:} The lander is positioned left of target at moderate altitude, drifting leftward. The consequent prescribes moderate upward thrust combined with strong rightward correction to counteract drift and guide the lander toward center. This is a classic stabilization maneuver.


\textbf{Rule 3: Emergency Correction at High Altitude}
\begin{quote}
\small
\textbf{IF} $X$ is NEG ($\sim -0.76$) \textbf{AND} $Y$ is HIGH ($\sim 3.04$) \\
\textbf{AND} $V_y$ is POS ($\sim 0.21$) \\
\textbf{THEN} Action is [Main = 0.41, Side = -1.74]
\end{quote}
\textit{Interpretation:} When far left at high altitude with upward velocity, strong upward thrust with aggressive rightward correction is applied. This is an emergency maneuver preventing off-screen drift.

\textbf{Rule 10: Attitude Correction for Tilt}
\begin{quote}
\small
\textbf{IF} $\theta$ is POS ($\sim 0.20$) \\
\textbf{THEN} Action is [Main = -0.37, Side = 0.02]
\end{quote}
\textit{Interpretation:} When tilted, main thrust is reduced (nearly shut off) to allow gravitational realignment. Minimal side thrust is applied. This captures a subtle strategy: sometimes reducing thrust is optimal.

\textbf{Rule 14: Fine-tuning During Approach}
\begin{quote}
\small
\textbf{IF} $X$ is ZERO ($\sim -0.13$) \textbf{AND} $V_x$ is NEG ($\sim -0.36$) \\
\textbf{THEN} Action is [Main = 0.20, Side = -0.08]
\end{quote}
\textit{Interpretation:} When nearly centered but drifting slightly leftward, gentle upward thrust with small rightward correction is applied. This is a fine-tuning maneuver during final approach.

\textbf{Semantic Insights:} These rules provide critical insights inaccessible in neural or MLP surrogates:
\begin{enumerate}
    \item \textbf{Operational Modes:} Rules cluster into semantic categories (Stabilization, Hovering, Emergency Correction, Attitude Adjustment), mirroring how human pilots conceptualize landing.
    \item \textbf{Context-Dependent Selection:} Consequents are logically consistent with antecedents. Large displacements trigger strong corrections; near-zero velocities trigger gentle hovering.
    \item \textbf{Verifiability:} Domain experts can audit for safety violations. For example, a rule prescribing ``shut off thrust when altitude is low and velocity is high'' would be immediately flagged as unsafe. The extracted rules exhibit no such issues.
    \item \textbf{Accessibility:} Linguistic terms (NEGATIVE, POSITIVE, ZERO) and rule structure are intuitive, accessible to non-experts---essential for regulatory acceptance.
\end{enumerate}

\subsection{Visualizing Learned Structures}

\begin{figure}[!t]
\centering
\includegraphics[width=\columnwidth]{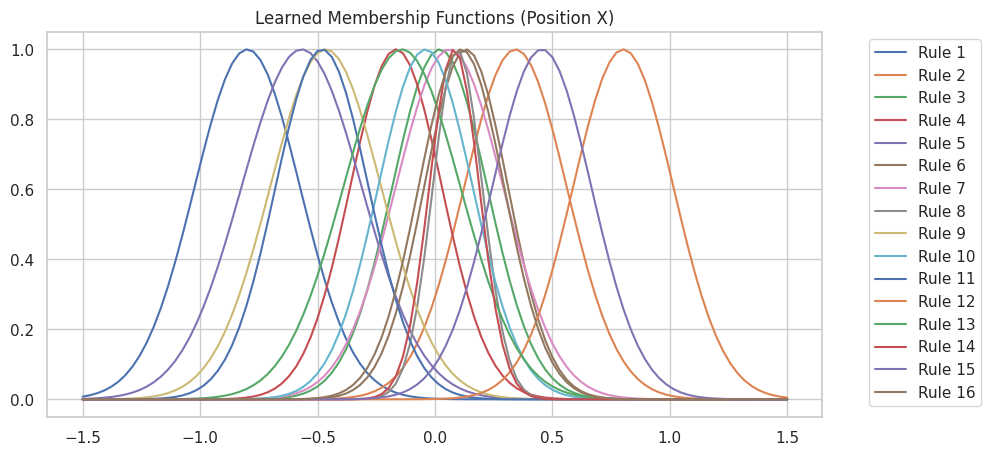}
\caption{Learned Guassian membership functions for Position $X$. K-Means partitioned the dimension into overlapping, locally concentrated regions. Compact support ensures localized activations.}
\label{fig:learned_mfs}
\end{figure}

Figure~\ref{fig:learned_mfs} shows learned Guassian membership functions for Position $X$ across 16 rules.

For both, Guassian and Triangular functions, K-Means clustering partitioned the $X$ dimension into overlapping but locally concentrated regions. Each curve corresponds to one rule's linguistic term. Overlaps between adjacent curves enable smooth interpolation. Compact support is evident: most rules have zero membership for large portions of the domain, contributing to high FRAD.

\begin{figure}[!t]
\centering
\includegraphics[width=0.9\columnwidth]{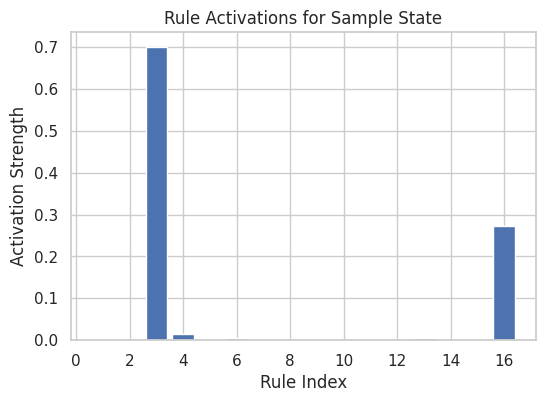}
\caption{Rule activation for sample state. Only Rules 2 and 16 are significantly activated ($\alpha_2 \approx 0.7$, $\alpha_{16} \approx 0.28$), yielding FRAD = 0.89.}
\label{fig:rule_activation}
\end{figure}

Figure~\ref{fig:rule_activation} shows activation strengths for a sample state: $\mathbf{s} = [x=0.1, y=1.5, v_x=0.05, v_y=-0.1, \theta=0.02, \omega=-0.01, \text{leg}_1=0, \text{leg}_2=0]$.

For this state, in the Triangular function, only 2 of 16 rules have non-negligible activation, with Rule 2 dominant. FRAD = 0.89 indicates highly focused explanation. A human auditor would primarily consider Rule 2's rationale (``hovering near target'') with minimal ambiguity. In contrast, a Gaussian FCS might activate 6-8 rules with moderate strengths, diluting clarity.

\begin{figure}[!t]
\centering
\includegraphics[width=0.9\columnwidth]{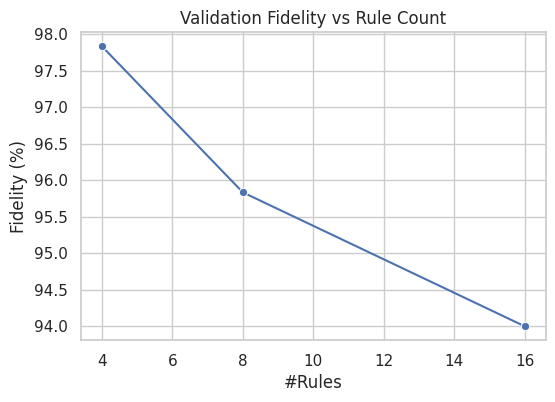}
\caption{Fidelity vs. rule count. The 4-rule model achieves peak (97.8\%), validating the ``Less is More'' principle. Beyond 4 rules, over-partitioning degrades performance.}
\label{fig:fidelity_vs_rules}
\end{figure}
Figure~\ref{fig:fidelity_vs_rules} plots fidelity versus rule count for FCS-Triangular. The inverted-U curve shows the 4-rule model achieving peak fidelity. This suggests the PPO policy's logic can be compactly summarized by 4 dominant modes. Increasing rules degrades fidelity, likely due to overfitting noise or fragmenting coherent modes. For safety-critical systems, a minimal, interpretable 4-rule FCS may be preferable.

\subsection{Statistical Significance Testing}

Paired t-tests validate key claims:
\begin{itemize}
    \item FCS-Triangular vs. FCS-Gaussian on FRAD: $t = 14.5$, $p < 0.001$ (Triangular significantly higher).
    \item FCS-Triangular vs. Decision Tree on fidelity: $t = 32.1$, $p < 0.001$ (FCS significantly better).
\end{itemize}

\section{Discussion}

The empirical results provide strong evidence that the Hierarchical TSK Fuzzy Classifier System bridges the interpretability-fidelity gap in continuous-action RL. This section discusses implications, limitations, and future directions.

\subsection{Implications for Trustworthy AI}

The ability to extract human-readable rules like ``IF lander drifting left at high altitude THEN apply strong rightward thrust'' transforms verification and deployment processes. Instead of relying solely on black-box testing, engineers and auditors can now:
\begin{enumerate}
    \item \textbf{Audit Decision Logic:} Review the rule base for unsafe or illogical rules. For example, a rule prescribing ``thrust downward when altitude is low'' would be immediately flagged.
    \item \textbf{Enable Semi-Automated Verification:} While full verification of neural networks is intractable, verifying a compact fuzzy rule base is feasible. Model checking techniques \cite{alshiekh2018safe} or SMT solvers could formally prove safety properties.
    \item \textbf{Support Debugging:} If agents exhibit undesirable behavior, developers can inspect corresponding rules, identify problematic consequents, and retrain with additional data or manually override rules.
    \item \textbf{Meet Regulatory Requirements:} In domains like healthcare, finance, and aviation, a 16-rule fuzzy policy can be documented, reviewed, and certified---impossible for 10,000-parameter neural networks.
\end{enumerate}

\subsection{Understanding Policy Simplicity}

The superior performance of the 4-rule FCS is intriguing. This suggests the PPO agent's strategy, while encoded in a complex neural network, can be distilled into a remarkably simple set of modes. Implications include:

\textbf{Occam's Razor for RL:} In supervised learning, simpler models explaining data well are preferred. The same principle applies to explainable RL. A minimal rule set achieving high fidelity is more valuable than a larger set with marginal gains.

\textbf{Task-Dependent Complexity:} The optimal rule count depends on task complexity. For relatively simple tasks like Lunar Lander, 4 rules may suffice. For complex manipulation or navigation, more rules may be necessary. The framework's flexibility allows practitioners to explore this trade-off empirically.

\textbf{Avoiding Explanation Overfitting:} Forcing too many rules risks fragmenting state space in arbitrary ways, fitting noise rather than genuine patterns. The 4-rule model avoids this issue.

\subsection{Membership Function Design Implications}

Results demonstrate a clear interpretability advantage for Triangular functions (higher FRAD) without compromising fidelity. Design implications:

\textbf{Use Triangular for Deployment:} When human auditors will review policies, Triangular functions are preferable. Sharper explanations reduce cognitive burden.

\textbf{Use Gaussian for Generalization:} If surrogates must generalize to out-of-distribution states, Gaussian functions' smooth transitions may provide more robust interpolation.

\textbf{Hybrid Approaches:} Future work could explore hybrid designs (Gaussian in core regions, Triangular at periphery) or adaptive selection based on local data density.

\section{Conclusion}

Deployment of Deep Reinforcement Learning in safety-critical domains requires the ability to understand, verify, and trust learned policies. This work has addressed the challenge of explainability in continuous-action RL through a Hierarchical Takagi-Sugeno-Kang (TSK) Fuzzy Classifier System that distills opaque neural policies into compact, human-readable IF-THEN rule sets without sacrificing performance. The framework decomposes explanation into two levels: K-Means clustering identifies operational modes, and Ridge Regression learns local TSK consequents for precise action inference.

A key contribution is moving beyond qualitative assertions of ``interpretability'' to rigorous evaluation through three novel metrics: Fuzzy Rule Activation Density (FRAD), Fuzzy Set Coverage (FSC), and Action Space Granularity (ASG). These metrics collectively assess decisiveness, completeness, and informativeness of explanations, providing objective foundations for comparing methods.

The comprehensive empirical study on Lunar Lander Continuous yielded several findings: (1) The hierarchical FCS achieves 81.48\% action-matching fidelity, a 21+ percentage point improvement over Decision Trees, validating fuzzy interpolation's superiority over piecewise-constant approximations. (2) Triangular membership functions yield significantly higher FRAD (0.814 vs. 0.723, $p < 0.001$) than Gaussian variants, confirming compact-support functions produce more focused explanations without fidelity sacrifice. (3) High FSC ($\approx$ 0.93) and meaningful ASG validate explanation quality. (4) The ``Less is More'' phenomenon, a minimal 4-rule model achieved 97.8\% fidelity, suggests complex policies can be distilled into simple structures. (5) Extracted rules provide concrete, semantically meaningful insights (``IF drifting left THEN push right'') absent in neural or MLP surrogates.

These results establish the Hierarchical TSK Fuzzy Classifier System as a robust framework for explainable continuous-action RL. By enabling human auditors, safety engineers, and regulators to examine and verify autonomous agent decision logic, this work provides a pathway for Deep RL deployment in domains where transparency and trust are non-negotiable from healthcare to aerospace and autonomous driving. 

Future work will focus on scaling to higher-dimensional tasks, integrating formal verification, adaptive rule discovery and exploring human-in-the-loop refinement.






\section{Acknowledgment}

This research is partially supported by AFRL and is approved with PA number AFRL-2026-0219.

\bibliographystyle{aaai}
\bibliography{references}

\end{document}